\documentclass[11pt,a4paper]{article}
\usepackage[hyperref]{acl2020}
\usepackage{times}
\usepackage{latexsym}

\usepackage{microtype}

\usepackage{amsmath}
\usepackage{graphicx}
\usepackage{soul}
\usepackage[ruled]{algorithm2e} 

\usepackage{booktabs}
\usepackage{CJKutf8}
\newcommand{\specialcell}[1]{%
  \begin{CJK*}{UTF8}{gbsn}#1\end{CJK*}}

\aclfinalcopy 
\def\aclpaperid{23X-P8A4E8F3B3} 

\title{A Hierarchical Location Normalization System for Text}

\author{
	Dongyun Liang, Guohua Wang, Jing Nie, Binxu Zhai and Xiusen Gu \\
	Tencent, China\\
	{\tt \{dylanliang, guohuawang, dennisgu\}@tencent.com}
}

\date{}

\begin{document}
\maketitle
\begin{abstract}
It's natural these days for people to know the local events from massive documents.
Many texts contain location information, such as city name or road name, which is always incomplete or latent.
It's significant to extract the administrative area of the text and organize the hierarchy of area, called location normalization.
Existing detecting location systems either exclude hierarchical normalization or present only a few specific regions.
We propose a system named ROIBase~\footnote{\small \url{github.com/waterblas/ROIBase-lite}} that normalizes the text by the Chinese hierarchical administrative divisions.
ROIBase adopts a co-occurrence constraint as the basic framework to score the hit of the administrative area,
achieves the inference by special embeddings, and expands the recall by the ROI (region of interest).
It has high efficiency and interpretability because it mainly establishes on the definite knowledge and has less complex logic than the supervised models.
We demonstrate that ROIBase achieves better performance against feasible solutions and is useful as a strong support system for location normalization.
\end{abstract}

\section{Introduction}
In every day and every place, various events are being reported in the form of texts, 
and many of these don't present hierarchical and standard locations.
In the context-aware text, location is a fundamental component that supports a wide range of applications.
We need to focus on the normalizing location to process massive texts effectively in specific scenarios.
As the text stream in social media are more quickly in accident or disaster response~\cite{munro2011subword},
location normalization is crucial for situational awareness in these fields, 
in which the omitted writing style often avoids redundant content.
For example, ``\specialcell{十陵立交路段交通拥堵} (Traffic congestion at Shiling Interchange)'' refers to a definite location, but there's no indication of where the Shiling Interchange is to make an exact response,
unless we know it belongs to Longquanyi district, Chengdu city, Sichuan province.

Countries are divided up into different units to manage their land and the affairs of their people easier.
Administrative division (AD) is a portion of a country or other region delineated for the purpose of administration.
Due to China's large population and area, the AD of China have consisted of several levels since ancient times.
For clarity and convenience, we cover three levels in our system, and treat the largest administrative division of a country as 1st-level, next subdivisions as 2nd-level and 3rd-level, 
which matches the provincial (province, autonomous region, municipality, and special administrative region), prefecture-level city and county in China, shown as Table~\ref{fig:administrative divisions ch}.
China administers more than 3,200 divisions in these flattened levels.
In such a large and complex hierarchy, much work stops at extracting the relevant locations, such as named entity tagging~\cite{srihari2000hybrid}. 
There are many similar named entity recognition (NER) toolkits~\cite{che2010ltp, finkel2005incorporating} for location extraction.
As the ambiguity is very high for location name,
~\citet{li2002location} and \citet{al2017location} develop to the disambiguation of location extraction.
We take a step closer to extract normalization information, and determine which the three hierarchical administrative area the document mainly describes.

\begin{table*}
\centering
  \includegraphics[width=12cm,height=6cm]{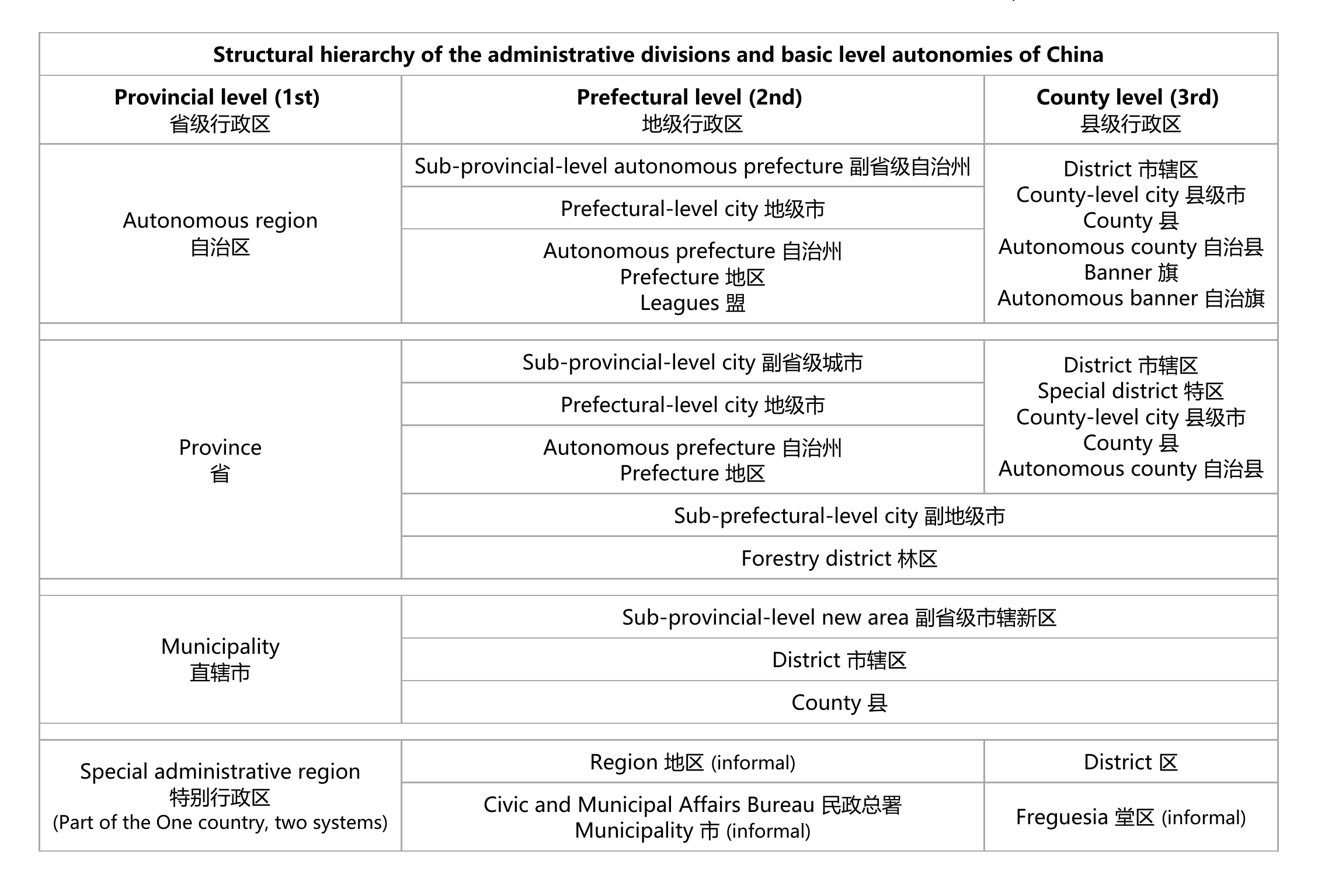}
  \caption{Structural hierarchy of the administrative divisions.}
  \label{fig:administrative divisions ch}
\end{table*}

The challenges are a bit different in our location normalization, which are mainly in ambiguity and explicit absence.
For example, there is a duplicate Chaoyang district as 3rd-level in Beijing and Changchun city, and ``Chaoyan'' also means the rising sun in Chinese, which may cause ambiguity.
If ``Beijing'' and ``Chaoyang'' are mentioned in the same context, 
it is confident that ``Chaoyang'' should refer to the district of Beijing city.
Similarly, \citet{yarowsky1995unsupervised} proposes a corpus-based unsupervised approach that avoids the need for costly truthed training data.
However, it's common that some contexts lack enough co-occurrence of AD to disambiguate or the explicit information completely misses.
We refer to it as the explicit absence problem, and neither NER nor disambiguation makes it work unless more hidden information is explored.
There are many specific AD-related points identifying which division is, including:

\begin{itemize}
\setlength{\itemsep}{1pt}
	\item Location alias, e.g. ``\specialcell{鹏城} (Pengcheng)'' is the alias name of Shenzhen city;
	\item Old calling or customary title, e.g. ``\specialcell{老闸北} (Old Zhabei)'' is a municipal district that once existed in Shanghai city;
	\item The phrase about the spatial region event, e.g. ``\specialcell{中国国际徽商大会} (China Huishang Conference)'' has been held in Hefei city;
	\item Some POIs (point of interest), e.g. The well-known ``\specialcell{颐和园} (Summer Palace)'' is situated in the northwestern suburbs of Beijing.
\end{itemize}

We summary them as a concept named ROI, which is both similar and different from POI.
POI dataset collects specific location points that someone may find useful or interesting. 
It maps the detailed address that covers the administrative division.
However, many POIs only build an uni-directional association with AD.
For example, Bank of China as a common POI is opened across the China.
We can find many Bank of China at a specific AD, but if only ``Bank of China'' exists in a context, we can't directly confirm its location without more area information.
Since POI is uncertain naturally,
we propose the concept of ROI, which has a bi-directional association with AD.
Given an ROI mapping the fixed hierarchical administrative area, ROI has high confidence to represent the area,
as well as the area contains it definitely.
In the absence of explicit patterns, the co-occurring ROI in the context can be good evidence to predict the most likely administrative area.
The main contributions of the system are as follows, which can be applied to other languages:
\begin{figure*}
  \centering
  \includegraphics[width=16cm,height=4cm]{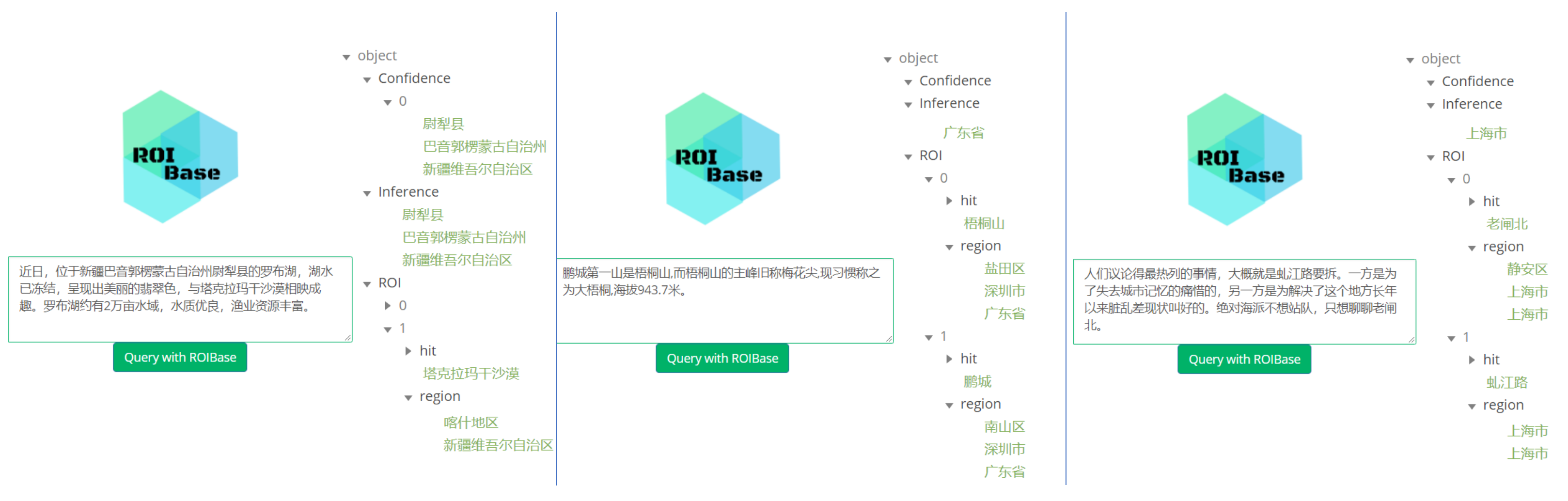}
  \caption{User interface of ROIBase.}
  \label{fig:demo1}
\end{figure*}

\begin{enumerate}
\setlength{\itemsep}{1pt}
	\item We provide a structured AD database, and use the co-occurrence constraints to make a decision;
	\item The ROIBase is equipped with geographic embeddings trained by special location sequences to make an inference;
	\item We use a large news corpus to build a knowledge base that is made up of ROIs, which helps normalization.
\end{enumerate}

\section{User Interface}

We design a web-based online demo~\footnote{\url{http://research.dylra.com/v/roibase/}} to show the location normalization.
As shown in Figure~\ref{fig:demo1}, there are three cases split by blue lines, and each case mainly contains two components: query and result.

\textbf{Query} \ Input the document into the textbox with a green border to query for ROIBase. The query accepts the Chinese format sentences, such as the text from news or social media.

\textbf{Result} \ On the right of the textbox, it will show the structured result from ROIBase after submitting the query.
The result consists of there parts: \textit{Confidence}, \textit{Inference} and \textit{ROI}.

\textit{Confidence} represents the result that can be extracted and identified from explicit information.
For example, we have confidence to fill ``\specialcell{新疆} (Xinjiang)'' when ``\specialcell{尉犁县} (Yuli County)'' and ``\specialcell{巴音郭楞蒙古自治州} (Bayingol Mongolian Autonomous Prefecture)'' are coming together in context.

\textit{Inference} is complement for the \textit{Confidence} by embeddings, where the nearest uncertain administrative level will be inferred from the implicit information of the input.
For example, none of the explicit administrative area appears in middle case of the Figure~\ref{fig:demo1}, so the \textit{Inference} will start with 1st-level (the largest division), and it infers ``\specialcell{广东省} (Guangdong Province)''. 
If the \textit{Confidence} comes up with 1st-level, the \textit{Inference} will start with 2nd-level.
If the \textit{Confidence} is filled with three levels, \textit{Inference} does nothing and keeps it as before.

\textit{ROI} is derived from the ROI knowledge base. 
We will match the input with the ROI knowledge base, and return the ROI associated with the administrative area when the match is successful.
The types of ROI are many and varied, and what they have in common is that it build the bidirectional relation with a hierarchical AD.
As shown in Figure~\ref{fig:demo1}, ``\specialcell{梧桐山} (Wutong Mountain)'', the highest peak in Shenzhen city,  map to three levels: [Yantian district, Shenzhen city, Guangdong province].

When the user queries, the input will be segmented into tokens by a Chinese tokenizer.
Two processes are running in parallel: one is calculating the \textit{Confidence} and then \textit{Inference}, the other is retrieving the ROI knowledge base.
The final result will be restructured back to the front in green color.

\section{Approach}

\subsection{Administrative Division Co-occurrence Constraint}
We support an administrative division database, including the names and partial aliases of the administrative areas in China, which are organized in the form of hierarchy.
Each record is associated with its parent and children, for example, ``\specialcell{襄阳市} (Xiangyang city)'' is at 2nd-level, its alias is Xiangfan, its parent is Hubei province, and some children of its divisions are Gucheng County, Xiangzhou District, etc.
we develop a co-occurrence constraint based on this database to \textit{Confidence} result, shown in Algorithm~\ref{alg:generator}.

\begin{algorithm}[htb]
	\caption{processing \textit{Confidence}}
	\label{alg:generator}
	 \textbf{Input}: S, sentences from text  \\
     \textbf{Output} D, hierarchical administrative division\\
     $ T \leftarrow \emptyset, \  Q \leftarrow \{\} $ \\
	\ForEach{{\rm word phrase} $w \in S $}{
        \uIf{$w$ hit AD database}{
            expand $w$ to three levels $[l_1,l_2,l_3]$ by standard AD, and add them into $T$\;
        }
	}
	\ForEach{{\rm hierarchical candidate} $t \in T $}{
        Count the hit number of level of $t$ in $S$: \\
        $Q[t] = CountLevel(S, t)$ \\
	}
    Filter out $t \in Q $ when $Q[t] < \max{Q}$ \\
	\ForEach{ {\rm filtered} $t \in Q $ }{
	 	\ForEach{ {\rm sentence} $s \in S $ }{
            $Q[t] += \frac{Count(s, t)}{(1+ CountOtherAD(s))}$  \\
	    }
	}
	\Return $D= \mathop{\arg\min}_{t[:\max{Q}]}(Q[t])$

\end{algorithm}

Firstly, we expand the possible AD hierarchy as candidates based on the input segments, and filter the longest to next calculation.
If a sentence is full of various AD information, it is probably just the listing of addresses that makes no sense, such as:

{\small \specialcell{青少年橄榄球天行联赛总决赛在\underline{上海森兰体育公园}举行。 由来自\underline{北京}、\underline{上海}、\underline{深圳}、\underline{重庆}、\underline{贵阳}等地的青少年选手组成的...}} \\
where the underlined words are related to the administrative area.
The more various area-related words are, and the less certainty a sentence has.
We consider the frequency of the hits as well as the penalty of other surrounding area-related words,
and construct a function to accumulate the weight of each sentence for AD.
Finally, we get the \textit{Confidence} result based the explicit statistics.

\subsection{Geographic Embeddings}\label{Geographic Embeddings}
We propose to train geographic embeddings by word sequences related to AD.
As the location information in a document is usually only a small part,
the standard name of AD are sparse and disperse, and the words related to geographic locations (now called geographic words) in a long tail are rarely seen.
We don't directly get the embedding from the raw word sequences, and instead, 
we assume that the raw sequences are made up of the records of AD database, geographic words, and others.
To keep the former twos, we pass through a large news corpus, more than 14.3 million documents, take every phrase of news sentences that hits the AD records as a starting point,
use a NER toolkit to recognise the location entities among the surrounding two sentences, and keep order to extract the candidate sequences that consist of the standard AD records and location entities.
In the pattern of the NER model, it's not extremely accurate, and various types of the phrases related to location are generically recognized.
We collect the candidate sequences greater than a threshold length to train geographic embeddings.

Given a set $S$ of candidate sequences extracted from documents,
each sequence $s = (w_1, ..., w_m ) \in S $ is made up of the AD records and location entities,
where the relative order of elements in $s$ stays the same as raw text.
The aim is to learn a $d$-dimensional real-valued embedding $v_{w_i}$ of each $v_{w_i}$, 
so that the administrative area and geographic words are in the same embedding space,
and the adjacent administrative areas lie nearby in the embedding space.
We learn the embedding using the skip-gram model~\cite{mikolov2013distributed} by maximizing the objective function $\mathcal L$ over the set $S$,
which is defined as follows:
$$
\mathcal{L} = \sum\limits_{s \in S} \sum\limits_{w_i \in s} (\sum\limits_{-n \geq j \leq n, i} \log P(w_{i+j} | w_i)) 
$$
$$
P(w_{i+j} | w_i) = \frac{\exp(v_{w_i}^T v_{w_{i+j}}^{'})}{\sum_{w=1}^{|V|} \exp(v_{w_i}^T v_{w}^{'})}
$$
where $v$ and $v^{'}$ are the input and output vector, $n$ is the size of the sequence window, and $V$ is the vocabulary that consists of the administrative areas and geographic words. 

\begin{figure}[htb]
	\includegraphics[width=20em]{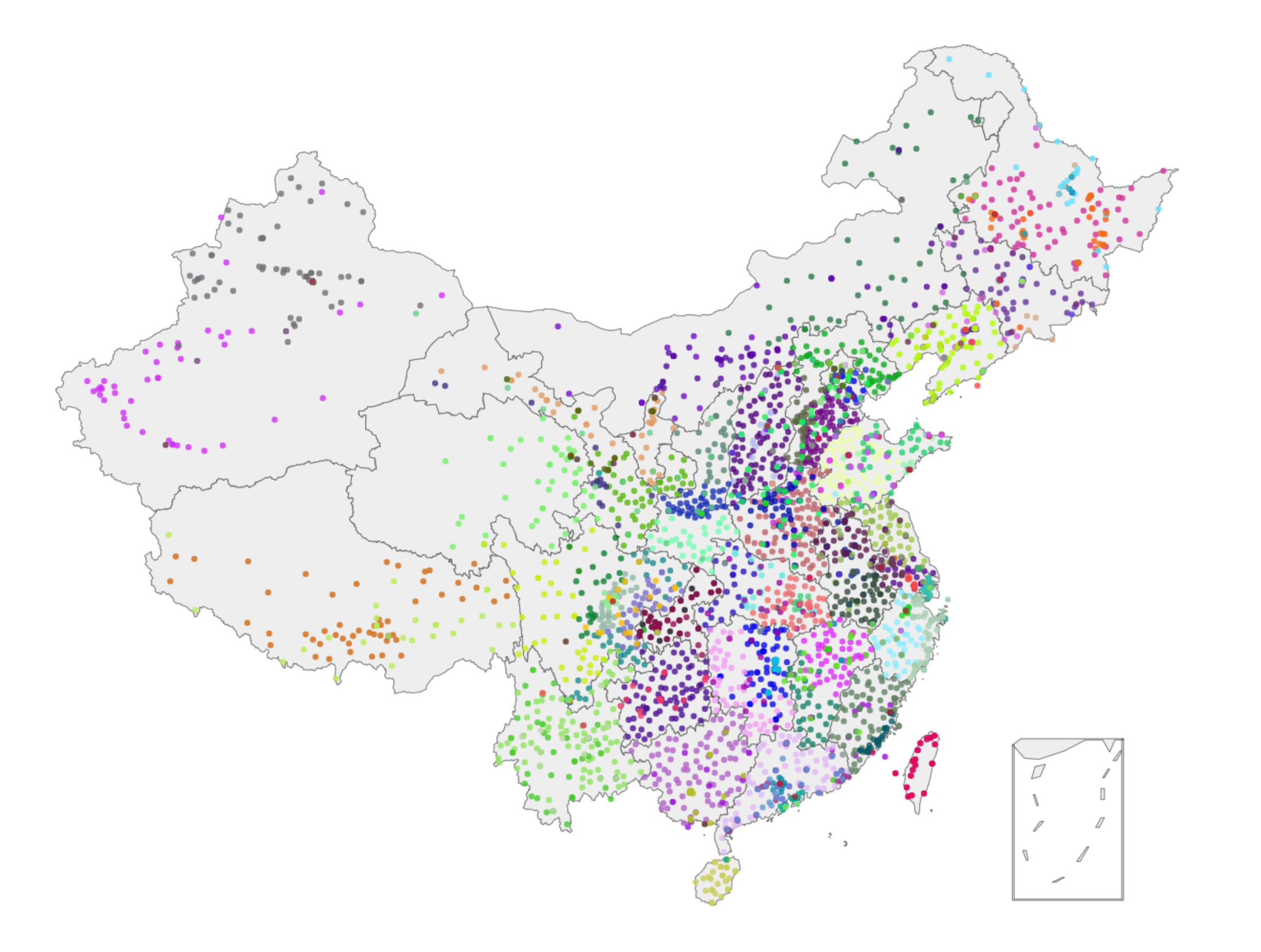}
	\caption{The clustering distribution of geographic embeddings about administrative areas.}
	\label{fig:embedding_cluster}
\end{figure}

To evaluate whether the region characteristics are captured by geographic embeddings,
we design a visualization to show.
Firstly, we perform k-means clustering on the learned embeddings of records in AD database,
cluster 4,000+ standard AD to 100 clusters, 
and then plot the scatters on the map of China with the division borders, where the different colors represent different clusters and the coordinates are the rough locations of the self standard AD. 
As shown in Figure~\ref{fig:embedding_cluster}, the scatters in same clusters are mainly located in same administrative area, 
and it means that the geographic similarity is well encoded.

\subsection{Inference}
Based on the \textit{Confidence} result, 
we utilize the geographic embeddings that we train in the above section to infer the next administrative area.
We first get the intersection of the input text and the geographic words $V$, and average the embeddings of the intersection at each dimension as the representation of the input $v_{input}$.
Then we embed the latest level's divisions of \textit{Confidence} to get the candidate embeddings.
For example, the \textit{Confidence} ends with 2nd level, denoted as $[l_1,l_2]$, 
so the embeddings of its latest level's divisions $[l_2^1, ...l_2^k]$ can be denoted as $v_{l_2^i}, i=[1,...,k]$, where $k$ is the number of $l_2$ subdivisions. 
It can be observed that cosine similarities between the right candidate and geographic embedding are often higher compared to other candidates embeddings.
We make the \textit{Inference} by $\arg\max_{l_2^i}Cosine(v_{l_2^i}, v_{input})$ as the complement of \textit{Confidence}.

\subsection{ROI}
Since embeddings are implicit, we build an ROI knowledge base to improve interpretability and reduce the bias of \textit{Inference}.
Unlike traditional taxonomies that require a lot of manual labor, we propose a novel method to extract ROI from large corpus, 
which uses the statistics to model inconsistent, ambiguous and uncertain information it contains.

Given the geographic sequences $s$ in section \ref{Geographic Embeddings}, where $\bar{w} \in s$ is the geographic word,
we assume that the most frequent administrative area in the window of the geographic word probably corresponds to its division.
In fact, some administrative area records appear more frequently in general,
such as Beijing, Shanghai and other big cites.
We consider the number of the pair $(\bar{w}, w_i)$ appears in the $S$, where $w_i$ represents the administrative area name. 
and offset by the total count of $w_i$ in the whole corpus.
Therefore, a similar tf–idf weighting scheme is applied to balance the exact division:
$$
	score(\bar{w}, w_i) = Count(\bar{w}, w_i) \times IDF(w_i) 
$$
where the $Count$ denotes the counting operation of the co-occurrence of $\bar{w}$ and $w_i$ in each geographic sequence,
and $IDF$ denotes the inverse document frequency of $w_i$ in all sequences for $S$.

We score each pair $\bar{w}$ and $w_i$, and filter the valid pairs by a high threshold.
Then the sorted mapping $ \{\bar{w} | (w_1, g_1), ...,(w_t, g_t)\} $ is obtained for each $\bar{w}$, where $g_i$ denotes the score weight,
the higher $g_i$ ranks more ahead.
It is noteworthy that the geographic word is not equal to ROI.
We use the information entropy to filter the valid candidates:
$$
E(\bar{w}) = -\sum\limits_{i}P_{i}\log P_i, \ P_i=\frac{g_i}{\sum_{j}^{t}g_j}
$$
If $\bar{w}$ can't represents the administrative area, the weights of the candidates mappings will be dispersed.
The higher $E(\bar{w})$ is, the less certain the the mapping contains.
We cut off the high $E(\bar{w})$ to keep the candidates of ROIs.

For a specific candidate ROI, it is common that the upper level of mapping will has the higher frequency than the low level in news corpus.
For example, the co-occurrence of \textit{Summer Palace} and \textit{Beijing} is larger than the co-occurrence of \textit{Summer Palace} and \textit{Haidian},
and Haidian district is a subdivision of Beijing city. 
We base subdivision relation to correct the weight of $w_i$ when the $w_{j}$ is the parent division of $w_i$, where $j<i$.
$$
g_i = g_i / P(w_i|w_j, \neg w_i, s)
$$
$$
P(w_i|w_j, \neg w_i, s) = \frac{\sum_{s \in S} H(w_i \cap f(s))}{\sum_{s \in S} H(w_j\cap \neg w_i \cap s)}
$$
where $\neg w_i$ means the operation without $w_i$, $P(w_i|w_j, \neg w_i, s)$ denotes the probability that only $w_j$ appears in $s$ but actually it belong to $w_i$,
$f(s)$ denotes the sequences that are in the same document excluding $s$,
and $H$ is the Heaviside step function.

We sort the mapping again by the re-weight scheme, 
and get the top few pairs, which are on same orders of magnitude, to compose ROI pairs $(\bar{w}, <l_1, l_2, l_3>)$, where $l_1, l_2, l_3$ represent the three levels of AD and it will be set to null if one is missing.
Finally, the pairs are inserted into Elasticsearch~\footnote{\url{https://www.elastic.co}} engine to build the knowledge base.
\begin{table*}[ht]
    \centering
    \begin{tabular}{p{0.15\textwidth}p{0.15\textwidth}p{0.6\textwidth}}
    \toprule
    ROIBase & NER+pattern & section of text \\
    \hline
    \textbf{news} \\
	\hline
    \small{\specialcell{-,呼和浩特市,内蒙古自治区}}  & \small{\specialcell{内蒙古}} & \small{\specialcell{\underline{内蒙古大兴安岭}原始林区雷击火蔓延...}} (Lightning fire spreads in the virgin forest area of the Greater Xing'an Mountains, Inner Mongolia...) \\
    \small{\specialcell{-,深圳市,广东省}}  & 
    - & 
    \small{\specialcell{日前，\underline{华为基地}启用了无人机送餐业务...}} (A few days ago, Huawei base launched drone food delivery business...) \\
    \small{\specialcell{双流区,成都市,四川省}}  & 
    \small{\specialcell{海口}} & 
    \small{\specialcell{四川航空3u8751成都至海口航班...安全落地\underline{成都双流国际机场}...}} (Sichuan Airlines flight 3u8751 from Chengdu to Haikou returned and landed safely at Chengdu Shuangliu International Airport...) \\
	\hline
    \textbf{Weibo} \\
	\hline
	\small{\specialcell{-,丽江市,云南省}}  & 
    \small{\specialcell{-}} & 
    \small{\specialcell{拍不出\underline{泸沽湖}万分之一的美这个时节少了喧嚣多了闲适}} (Can't shoot one-tenth of the beauty of Lugu Lake...) \\
 	\hline
	\small{\specialcell{-,武汉市,湖北省}}  & 
    \small{\specialcell{湖北}} & 
    \small{\specialcell{\underline{湖北经济学院}学生爆料质疑校园联通宽带垄断性经营}} (Students from Hubei University of Economics questioned campus unicom's ...) \\   
    \end{tabular}
    \caption{The examples of the location extraction by ROIBase and NER}
    \label{tab:news}
\end{table*}

\section{Experiment}
There are no publicly available datasets on text location normalization, so as no comparable methods.
As many similar schemes about detecting location start from NER,
we build NER+pattern as baseline, which uses NER to recognise and retrieves the AD database.
We conduct the experiments on news and Weibo (social media in China) corpus.
The news contains title and content, the title is usually short and cohesive, 
and the content always has hundreds of words with more location information, of which the changes lie in redundancy and efficiency. 
The Weibo corpus is short-text, and the location information is always implicit.

We manually sample the finance and 
social news, and obtain 760 news that can be assigned to a definite place to build the news dataset.
Equally, 1228 short-texts are finally picked from Weibo corpus.
Location information is extracted by ROIBase and NER~\cite{che2010ltp}+pattern respectively on these datasets.
As the Table~\ref{tab:news} lists examples of the results,
only NER+pattern matching can't utilize the hidden information to completely normalize the locations,  
ROIBase contains 1.51 million geographic embeddings and 0.42 million ROIs, so it knows the more linking of AD by the underlined phrases.
\begin{table}[htp]
\centering
	\begin{tabular}{lcc}
		\toprule
		 &\multicolumn{2}{c}{F1-score}\\
		\textsc{method} & news & Weibo\\
		\hline
		ROIBase & 0.812 &  0.780\\
		NER+pattern & 0.525 & 0.582\\
		\hline
	\end{tabular}
	\caption{F1 score on two datasets}
	\label{tab:F1}
\end{table}

A variant of F1 score is used to measure the performance, which takes the incomplete output as 0.5 hit when counting.
As shown in Table~\ref{tab:F1}, ROIBase achieves better performance against NER with AD patterns by large margins.
Some of Weibo texts carry the label of location, and it contributes to the recognition of AD patterns, which closes the gap with us.
The long texts provide more abundant information, and ROIBase can eliminate confusion to improve the performance.

\begin{table}[htp]
\centering
	\begin{tabular}{ccccc}
		\toprule
		total & 1st & 2nd  & 3rd  & speed\\
		\hline
		36.8\% & 23\% & 48.7\% & 28.3\% & 751KB/s\\
		\hline
	\end{tabular}
	\caption{ROIBase statistics on 100,000 news}
	\label{tab:val}
\end{table}

Statistics over 100 thousand news from financial and social domains by ROIBase access to detailed results.
As shown in Table~\ref{tab:val}, we can normalize locations from 36.8 percent in general.
Among them, there is 23 percent normalization only at the 1st level, 48.7 percent at 2nd level,  and 28.3 percent with complete divisions.
We show the speed on a machine with Xeon 2.0GHz CPU and 4G Memory, 
and the speed of ROIBase is up to 751KB/s when the NER method~\cite{che2010ltp} costs 14.4KB/s.
ROIBase lets the user process vast amounts of long text in location normalization.

\section{Related Work}
\citet{zubiaga2017towards} makes use of eight tweet-inherent features for classification at the country level.
\citet{qian2017probabilistic} formalizes the inferring location of social media into a semi-supervised factor graph model,
and perform on the level of countries and provinces.
A hierarchical location
prediction neural network~\cite{huang-carley-2019-hierarchical} is presented for user geolocation on Twitter.
However, many of these focus on a single level, only cover fewer countries or states, or utilize extra features out of text.
There is room for improvement in the performance. 
Since \citet{mikolov2013distributed} proposes the word vector technique,
there are many applications.
\citet{grbovic2018real} introduces listing and user embeddings trained
on bookings to capture user’s real-time and long-term interest.
~\citet{wu2012probase} demonstrates that a taxonomy knowledge base can be constructed from the entire web in special patterns.
Inspired by the these cases, we make the first solution to normalize the location of text by hierarchical administrative areas.

\section{Conclusion}
Through the investigation, we found that there is very few work on location normalization of text, and the popular alike solutions, such as NER, are not directly transferable to it. 
The ROIBase system provides an efficient and interpretable solution to location normalization through a web interface,
which enables to process these modules with a cascaded mechanism.
We propose it as a baseline that can be applied in different languages easily, and look forward to more work on improving the location normalization.

\bibliography{anthology,acl2020}
\bibliographystyle{acl_natbib}

\end{document}